\documentclass[11pt]{article}
\usepackage{a4}
\usepackage{colacl}
 
\usepackage{epic}
\usepackage{eepic}
\usepackage{subfigure}
\usepackage{algorithmic}
\usepackage{graphics}
\newcommand{\s}[1]{\mbox{\emph{#1}}}
\title{\raisebox{0.5\baselineskip}[0in][0in]{\parbox{\linewidth}{\scriptsize\tt\begin{flushright}
        {\bf In:} Eisner, J., L. Karttunen and A.
        Th\'{e}riault (eds.), {\em Finite-State Phonology:\@
          Proc.\@ of the 5th Workshop \\ of the ACL Special
          Interest Group in Computational Phonology (SIGPHON)}, pp.\@
        57-67, Luxembourg, Aug.\@ 2000.
       \end{flushright}}} \\
       Taking Primitive Optimality Theory Beyond the Finite State}
\author{Daniel M. Albro\\Linguistics Department\\UCLA}
\date{\today}

\begin{document}
\thispagestyle{plain}
\pagestyle{plain}
\maketitle

\begin{abstract}
  Primitive Optimality Theory (OTP) \cite{Eisner97a,Albro98}, a
  computational model of Optimality Theory \cite{PrinceSmolensky93},
  employs a finite state machine to represent the set of active
  candidates at each stage of an Optimality Theoretic derivation, as
  well as weighted finite state machines to represent the constraints
  themselves.  For some purposes, however, it would be convenient if
  the set of candidates were limited by some set of criteria capable
  of being described only in a higher-level grammar formalism, such as
  a Context Free Grammar, a Context Sensitive Grammar, or a Multiple
  Context Free Grammar \cite{Seki91}.  Examples include reduplication
  and phrasal stress models.  Here we introduce a mechanism for
  OTP-like Optimality Theory in which the constraints remain weighted
  finite state machines, but sets of candidates are represented by
  higher-level grammars.  In particular, we use multiple context-free
  grammars to model reduplication in the manner of Correspondence
  Theory \cite{McCarthyPrince95}, and develop an extended version of
  the Earley Algorithm \cite{Earley70} to apply the constraints to a
  reduplicating candidate set.
\end{abstract}

\bibliographystyle{acl}

\section{Introduction}

The goals of this paper are as follows:
\begin{itemize}
\item To show how finite-state models of Optimality Theoretic
  phonology (such as OTP) can be extended to deal with non-finite
  state phenomena (such as reduplication) in a principled way.  
\item To provide an OTP treatment of reduplication using the standard
  Correspondence Theory account.
\item To extend the Earley chart parsing algorithm to multiple
  context free grammars (MCFGs). 
\end{itemize}

The basic idea of this approach is to begin
with a non-finite-state description of the space of acceptable
candidates (\emph{e.g.,} candidates with some sort of reduplication
inherent in them, or candidates which are the outputs of a syntactic
grammar), and to repeatedly intersect the high-level grammar
representing those candidates with finite state machines representing
constraints.  The intersection operation is one of weighted
intersection (where only the set of lowest-weighted candidates
survive) in order to model Optimality Theory, and will make use of a
modified version of the Earley parsing algorithm.

There are at least two alternative approaches to that which we will
propose here: to abandon finite state
models altogether and move to uniformly higher-level approaches
(\emph{e.g.,} \newcite{Tesar96}), or to modify finite state models
minimally to allow for (perhaps limited) reduplication (\emph{e.g.,}
\newcite{Walther00}).  The first of these alternative approaches deals
with context free grammars alone, so it would not be able to
model reduplicative effects.  Besides this, it seems preferable to
stick with finite-state approaches as far as possible, because
phonological effects beyond the finite state seem quite rare.  The
second of these approaches seems reasonable in itself, but it is not
suited for the type of analyses for which the
approach laid out here is designed.  In particular, Walther's approach
is tied to One-Level Phonology, a theory which limits itself to
surface-true generalizations, whereas the approach here is designed to
model Optimality Theory---a system with violable constraints---and in
particular Correspondence Theory.  Tesar's approach as well, while it
is a model of Optimality Theory, does not seem suited to
Correspondence Theory.  A final argument for using this approach, in
preference to one similar to Walther's approach, is that it can be
extended to cover other non-finite-state areas of phonology, such as
phrasal stress patterns, with no modification to the basic model.

\section{Quick Overview of OTP}

\subsection{Optimality Theory}

Optimality Theory (OT), of which OTP is a formalized computational model,
is structured as follows, with three components:
\begin{enumerate}
\item \textbf{Gen:} a procedure that produces infinite surface
  candidates from an underlying representation (UR)
\item \textbf{Con:} a set of constraints, defined as functions from
  representations to integers
\item \textbf{Eval:} an evaluation procedure that, in succession,
  winnows out the candidates produced by \textbf{Gen.}
\end{enumerate}
So OT is a theory that deals with potentially infinite sets of
phonological representations.  The OT framework does not by itself
specify the character of these representations, however.

\subsection{Primitive Optimality Theory (OTP)}

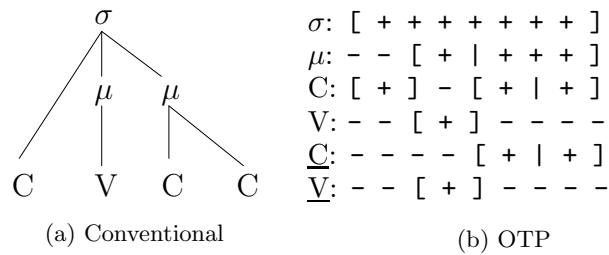
\begin{figure}
\begin{center} 
\begin{tabular}{cr} 
\subfigure[Conventional]{\setlength{\unitlength}{0.00087489in}
\begingroup\makeatletter\ifx\SetFigFont\undefined%
\gdef\SetFigFont#1#2#3#4#5{%
  \reset@font\fontsize{#1}{#2pt}%
  \fontfamily{#3}\fontseries{#4}\fontshape{#5}%
  \selectfont}%
\fi\endgroup%
{\renewcommand{\dashlinestretch}{30}
\begin{picture}(1473,1140)(0,-10)
\path(45,225)(540,990)
\path(540,990)(540,720)
\path(540,540)(540,180)
\path(540,990)(900,720)
\path(945,540)(945,225)
\path(945,540)(1395,180)
\put(900,0){\makebox(0,0)[lb]{\smash{{{\SetFigFont{12}{14.4}{\rmdefault}{\mddefault}{\updefault}C}}}}}
\put(1350,0){\makebox(0,0)[lb]{\smash{{{\SetFigFont{12}{14.4}{\rmdefault}{\mddefault}{\updefault}C}}}}}
\put(0,0){\makebox(0,0)[lb]{\smash{{{\SetFigFont{12}{14.4}{\rmdefault}{\mddefault}{\updefault}C}}}}}
\put(900,585){\makebox(0,0)[lb]{\smash{{{\SetFigFont{12}{14.4}{\rmdefault}{\mddefault}{\updefault}$\mu$}}}}}
\put(495,0){\makebox(0,0)[lb]{\smash{{{\SetFigFont{12}{14.4}{\rmdefault}{\mddefault}{\updefault}V}}}}}
\put(495,585){\makebox(0,0)[lb]{\smash{{{\SetFigFont{12}{14.4}{\rmdefault}{\mddefault}{\updefault}$\mu$}}}}}
\put(495,1035){\makebox(0,0)[lb]{\smash{{{\SetFigFont{12}{14.4}{\rmdefault}{\mddefault}{\updefault}$\sigma$}}}}}
\end{picture}
}} & 
\subfigure[OTP]{%
\begin{minipage}[b]{2in} 
$\sigma$: \texttt{[ + + + + + + + ]} \\ 
$\mu$: \texttt{- - [ + | + + + ]} 
\\ C: \texttt{[ + ] - [ + | + ]} \\ 
V: \texttt{- - [ + ] - - - -}\\ 
\underline{C}: \texttt{- - - - [ + | + ]} \\ 
\underline{V}: \texttt{- - [ + ] - - - -} 
\end{minipage} 
} 
\end{tabular} 
\end{center}

\caption{OTP Representation\label{fig:otp-rep}}
\end{figure}

The components of OT, as modeled by OTP (see
\newcite{Eisner97a}, \newcite{Eisner97b}, \newcite{Albro98}):
\begin{enumerate}
\item \textbf{Gen:} a procedure that produces from an underlying
  representation a finite state machine that represents all possible
  surface candidates that contain that UR (always an infinite set)
\item \textbf{Con:} a set of constraints definable in a restricted
  formalism---internally represented as Weighted Deterministic Finite
  Automata (WDFAs) which accept any string in the representational
  alphabet. The weights correspond to constraint violations. The
  weights passed through when accepting a string are the violations
  incurred by that string.
\item \textbf{Eval:} the following procedure, where $I$ represents the
  input FSM produced by Gen, and $M$ is a machine representing the
  output set of candidates:
  \begin{algorithmic}
    \STATE $M \gets I$
    \FORALL{$C_i\in \mathbf{Con}$, taken in rank order}
    \STATE $M \gets$ intersection of $M$ with $C_i$
    \STATE Remove non-optimal paths from $M$
    \STATE Zero out weights in $M$
    \ENDFOR
  \end{algorithmic}
\end{enumerate}
Representations in OTP are gestural scores using symbols from the set \(
\{-,+,[,],|\} \).  See Figure~\ref{fig:otp-rep} for an example.
This figure shows a CVCC syllable in a conventional notation, and also
in OTP notation.  The OTP notation is slightly more complex, though,
in that it also shows an underlying form for the syllable.  The
overlap relation of the conventional notation's association lines is
expressed in the OTP notation by the presence of constituent interiors
(``+'') in the same vertical slice through the diagram.  This
same-time-slice-membership relation is also used to show
correspondence.  Thus we see from this diagram that the surface
``CVCC'' syllable corresponds to underlying ``VCC,'' and that the
initial ``C'' does not correspond to any underlying segment.  Note
that tiers with no special marking are used to represent the surface
level of representation, and underlined tiers are used to represent
the underlying level of representation.

\section{Handling Reduplication: Overview}

\subsection{Overview}

Finite State Machines are useful in phonology because it is possible
to take any two finite state machines, each of which represents a set
of strings, and perform an \emph{intersection} operation on them.  The
resulting machine represents the intersection of the two sets of
strings.  For example, this allows us to use constraints represented
as FSMs to limit a candidate set.

Although we would sometimes like to characterize the candidate sets
using CFGs or MCFGs, it must be kept in mind that these formalisms do
not have the property of being intersectable with each other.  Thus,
in OTP terms, it would not be possible to represent the constraints as
CFGs or MCFGs.  However, there is a way out: it is possible to
intersect an FSM with a CFG or an MCFG.

Based on the above, an approach to handing reduplication in phonology
becomes clear---we start with an MCFG that enforces reduplicative
identity, then intersect it with the input FSM (produced by
\textbf{Gen}), then the constraint FSMs, as before.  The hard part,
then, is to come up with an efficient FSM-intersection algorithm for MCFGs
which also deals correctly with weighted FSMs.

\subsection{MCFGs}

A grammar formalism that is midway between CFGs and CSGs in expressive
power, an MCFG is like a CFG except that categories may rewrite to
tuples of strings instead of rewriting to just one string as usual. It
should be noted that MCFGs have been shown \cite{Vugt96} to be
equivalent to string-valued attribute grammars with only s-attributes,
relational grammars, and top-down tree-to-string transducers, so we could use any one of these grammars to provide a candidate space.  As an
example of an MCFG, here's a simple MCFG for the language \( \{ww|w\in
\{0,1\}^{+}\} \) (the language of total reduplication):
\[
\begin{array}{rcl}
S & \to & A_0 \: A_1 \\
A & \to & (1, 1) \\
  & | & (0, 0) \\
  & | & (0 \: A_0, 0 \: A_1) \\
  & | & (1 \: A_0, 1 \: A_1)
\end{array}
\]
The nonterminals of this grammar are \emph{S}, which has arity 1, and
\emph{A}, which has arity 2.  The right-hand sides of the productions
include notations such as $A_0$, which indicate the placement of each
part of the tuple-yield of any category.  Here, $A_0$ and $A_1$ are
the two parts of the single category $A$, so a rule like $A \to (0 \:
A_0, 0 \: A_1)$ indicates that $A$ rewrites to $0 \: 0 \: A$, with the
actual strings arranged in a tuple with a $0$ preceding the first part
of $A$ in the first half of the pair, and $0$ preceding the second
part of $A$ in the second half of the pair.

This grammar is in the normal form required by the algorithms presented here.  This normal form can be characterized as follows:
\begin{quote}
  For any category \( C \) of arity greater than 1, the category may
  appear in the right hand side of a production only if the right hand
  side refers to each element of \( C \) exactly once.
\end{quote}

\begin{figure}[tb]
{\par\centering \includegraphics{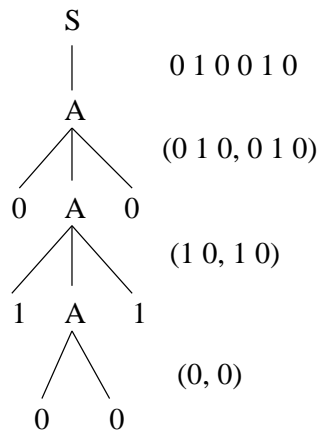}\par}
\caption{\label{fig:derivation}Derivation of ``010010''}
\end{figure}

A derivation of the string ``010010'' in this grammar would go as
follows: $S$ rewrites as $A_0 A_1$, that is, to the concatenation of
the string-yield of the two parts of $A$.  From here, $A_0$ and $A_1$
must both come from the parts of a single one of the four productions
for $A$.  $A$ then rewrites to $(0 \: A_0, 0 \: A_1)$, making, for
example, the value of $A_0$ in the $S$ production be $(0 \: A_0)$.
$A$ then rewrites to $(1 \: A_0, 1 \: A_1)$, so $S$ reduces to $0 
1 \: A_0 \: 0  1 \: A_1$.  Finally, $A$ rewrites to $(0, 0)$,
leaving the value of $S$ as $0 1 0 0 1 0$.  This derivation is
illustrated in Figure~\ref{fig:derivation}, the left side of which
depicts the derivation tree, while its right side shows (from the bottom
up) the string-yield of each non-terminal (shown just below and to the
right of it).

\subsection{Representation of Reduplicative Forms in OTP}

OTP constraints are inherently local---they can only refer to overlap
or non-overlap of interiors or edges in an instant of time.
Therefore, to enforce correspondences between forms, they must be
juxtaposed so as to occur in the same time-slices.  In OTP,
correspondence between the surface and underlying forms is established
by using one set of tiers for the surface form (each tier represents
either a feature or a type of prosodic constituent) and another
corresponding set for the underlying form.  For example, the tier
\emph{son} might specify the distribution of the surface feature
``sonorant'', while the tier \emph{\underline{son}} would specify its
underlying correspondent.  Elements of those tiers placed in the same
time-slice are considered to be in correspondence with one another.
In order to create correspondence between two portions of the same
surface form, then, we need to somehow have them simultaneously
juxtaposed so as to appear in the same segments of time and separated
in time as they will be on the surface.  This is accomplished by a
representational trick: in the example of reduplication, a copy of the
reduplicant's surface form is placed in a special set of tiers within
the base:
\begin{quote}
\begin{center}
\begin{tabular}{r|cc}
\textbf{SL}:& BASE & RED$_2$ \\
\textbf{UL}:& UR\( _{1} \)& UR\( _{2} \) \\
\textbf{RL}:& RED\( _{1} \)& --- \\
\end{tabular}
\vspace{1em}

--- or ---

\vspace{1em}
\begin{tabular}{r|cc}
\textbf{SL}:&  RED\( _{2} \) & BASE  \\
\textbf{UL}:&  UR\( _{2} \)    & UR\( _{1} \) \\
\textbf{RL}:&---&RED\( _{1} \) \\
\end{tabular}
\end{center}
\end{quote}
In these representations \textbf{SL} stands for the surface level of
representation, \textbf{UL} for the underlying level, and \textbf{RL}
for the special reduplicant level (the place where a copy of the
reduplicant is kept).  UR\( _{1} \) and UR\( _{2} \) are identical in
the input, and RED\( _{1} \) and RED\( _{2} \) need to be kept
identical by other means.  The means chosen here is to use an MCFG
enforcing the identity.  BASE-RED correspondence constraints operate
upon RED\( _{1} \) while templatic and general surface well-formedness
constraints operate upon RED\( _{2} \).  An example of this sort of
representation might help here.  Suppose that there are two surface
tiers, \emph{C} and \emph{V}.  Then a form such as [CV+CVC] (with CV
prefixing reduplication, assuming that the base is CVC, and with the
underlying form RED+/VC/) might be
represented as follows:
\begin{flushleft}
\begin{tabular}{rc}
C: &                         \texttt{[ + ] - - - [ + ] - [ + ]} \\
V: &                         \texttt{- - [ + ] - - - [ + ] - -} \\
\underline{C}: &             \texttt{- - - - [ + ] - - - [ + ]} \\
\underline{V}: &             \texttt{- - [ + ] - - - [ + ] - -} \\
\underline{\underline{C}}: & \texttt{- - - - - - [ + ] - - - -} \\
\underline{\underline{V}}: & \texttt{- - - - - - - - [ + ] - -} \\ 
INS: &                       \texttt{[ + ] - - - [ + ] - - - -} \\
DEL: &                       \texttt{- - - - [ + ] - - - - - -} \\
RDEL: &                      \texttt{- - - - - - - - - - [ + ]} \\
RED: &                       \texttt{[ + + + + + ] - - - - - -} \\
BASE: &                      \texttt{- - - - - - [ + + + + + ]} \\
\end{tabular}
\end{flushleft}
Note here that the special \emph{BASE} and \emph{RED} tiers indicate
the portions of the surface forms that are the base and reduplicant,
and that the reduplicant level of representation (that is, the level
that holds the copy of the reduplicant used for correspondence) is
present on the tiers labeled with double underlines.  The \emph{INS}
tier represents a time-discrepancy between the levels of
representation where time does not exist on the underlying level (so
the period of time taken up by the initial C in the surface
reduplicant and base doesn't correspond to anything in the underlying
level), and the \emph{DEL} tier represents time that does not exist
on the surface level, so the time taken up by the final C in the
underlying form of the reduplicant does not correspond to anything on
the surface.  The \emph{RDEL} tier is a mirror of the contents of
the \emph{DEL} tier in the surface reduplicant, and thus represents time
that does not exist in the special reference copy of the reduplicant.
This representation allows us to notice that the reduplicant fits a CV
template --- the left edge of it is aligned with a surface C, the
right edge with a surface V, and there are no other segments within
it.  (The relevant OTP constraints to reinforce this would be ``RED$[$
$\to$ C$[$,'' ``$]$RED $\to$ $]$V,'' ``$]$C $\bot$ C$[$ $\bot$ RED,'' and ``$]$C $\bot$
V$[$ $\bot$ RED,'' if highly ranked and in that order.)

\begin{figure*}
{\par\centering \includegraphics{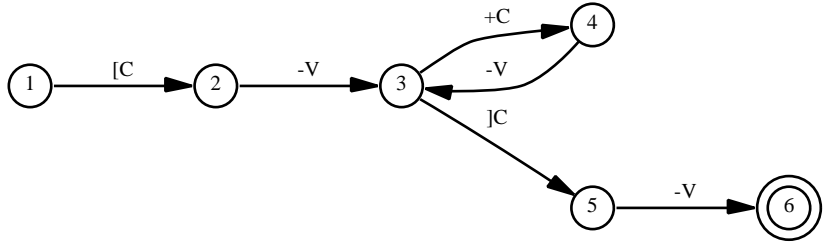} \par}

\caption{FSM Representation Used Here\label{fig:fsm1}}
\end{figure*}

In terms of
translating these representations to finite state machines (or to
strings), we use the alphabet $\{-,+,[,],|\}$, so that each FSM edge is
labeled with a member of this alphabet.  This representation differs
from that of earlier accounts of OTP, in that the FSM edges in those
accounts represented entire time slices, whereas an edge in this
representation represents a single tier in a time slice.  As an
example, the representation of:
\begin{quote}
\begin{tabular}{rccc}
C: & [ & +$^*$ & ] \\
V: & - & - & - \\
\end{tabular}
\end{quote}
is as shown in Figure~\ref{fig:fsm1}, where the ``C'' and ``V'' labels
are not part of the representation, but just there to ease reading.

\subsection{The Grammar Used}

The grammar used here is a bit complicated, but the important thing to
note about it is that it generates exactly the set of possible OTP
output forms in which the special reduplicant reference level of
representation contains an exact copy of the surface reduplicant,
placed within the time-duration of the base.  The grammar for a
situation in which there are two surface tiers appears in
Figure~\ref{fig:grammar}.  Extending this grammar to other numbers of
tiers is straightforward.  The constituents of this grammar are as
follows:
\begin{description}
\item[S] The start symbol.
\item[Non] Non-reduplicating material (such as non-reduplicating
  morphemes) before and/or after the reduplicating material.
\item[SSR] The surface tiers in a time-slice.
\item[UR] The underlying tiers in a time-slice.
\item[MRD] The reduplicant reference-level tiers in a time-slice where
  the tiers must contain the value $-$ (that is, outside of the base,
  which is the only place where the reduplicant level is used).
\item[Rd/Rd1/Rd2] The reduplicating part of an utterance.
\item[BDR] A right-facing boundary (allows anything to be in the
  surface tiers during its time-slice, and copies the right-facing
  half of that material into the reduplicant).
\item[BDL] A left-facing boundary (see $\s{BDR}$).
\item[B] The surface tiers in a time-slice plus identical material in
  the reduplicant tiers.  Thus $\s{B}$ represents an item in the
  reduplicant plus its copy in the special reduplicant reference
  level.
\end{description}
The remaining non-terminals define different values for the INS, DEL,
RDEL, RED, and BASE tiers, where INS and DEL are as defined in
\newcite{Albro98}, RDEL represents time that does not exist in the
reduplicant, RED represents the reduplicant (as a morpheme boundary),
and BASE represents the base as a morpheme boundary:
\begin{description}
\item[NBR] represents the state of not being in the base or the
  reduplicant.
\item[RLE] represents the left edge of the reduplicant.
\item[RRE] represents the right edge of the reduplicant.
\item[BLE] represents the left edge of the base.
\item[BRE] represents the right edge of the base.
\item[RB] represents a boundary between a reduplicant and a base,
  where the reduplicant comes first.
\item[BR] represents the reverse of $\s{RB}$.
\item[RED] represents the inside of the reduplicant.
\item[BASE] represents the inside of the base.
\end{description}
In this grammar any given time-slice will be defined as
$\s{SSR}$ or the first component of one of the $B$ categories,
followed by $\s{UR}$, followed by $\s{MRD}$ or the second component of
one of the $B$ categories, followed by one of the $\s{NBR}$, etc.,
categories.

\begin{figure}

\begin{footnotesize}
\[
\begin{array}{rcl}
S & \to & \s{Non} \: \s{Rd} \: \s{Non} \\
  & |  & \s{Rd} \: \s{Non} \\
  & |  & \s{Non} \: \s{Rd} \\
  & |  & \s{Rd} \\
\s{Non} & \to & \s{SSR} \: \s{UR} \: \s{MRD} \: \s{NBR} \\
        & |  & \s{Non} \: \s{SSR} \: \s{UR} \: \s{MRD} \: \s{NBR} \\
\s{SSR} & \to & A \: A \\
\s{UR} & \to & A \: A \\
\s{MRD} & \to & - \: - \\
\s{Rd} & \to & \s{Rd1}_0 \: \s{Rd1}_1 \\
\s{BDR} & \to & \left(
\begin{array}{l}
\s{BDR0}_0 \: \s{BDR1}_0 ,\\
\s{BDR0}_1 \: \s{BDR1}_1 \\
\end{array}
\right) \\
\s{BDL} & \to & \left(
\begin{array}{l}
\s{BDL0}_0 \: \s{BDL1}_0 ,\\
\s{BDL0}_1 \: \s{BDL1}_1 \\
\end{array}
\right) \\
\s{B} & \to & \left(
\begin{array}{l}
\s{B0}_0 \: \s{B1}_0 ,\\
\s{B0}_1 \: \s{B1}_1 \\
\end{array}
\right) \\
A & \to & - \; {\small |}\; + \; {\small |} \; [ \; {\small |} \; ] \; {\small |} \; {\tiny |} \\
\s{BDR}_n & \to & \left(
\begin{array}{l}
-, \\
-
\end{array}
\right|
\left.
\begin{array}{l}
+, \\
+
\end{array}
\right|
\left.
\begin{array}{l}
{[}, \\
{[}
\end{array}
\right|
\left.
\begin{array}{l}
{]}, \\
-
\end{array}
\right|
\left.
\begin{array}{l}
|, \\
{[}
\end{array}
\right) \\
\s{BDL}_n & \to & \left(
\begin{array}{l}
-, \\
-
\end{array}
\right|
\left.
\begin{array}{l}
+, \\
+
\end{array}
\right|
\left.
\begin{array}{l}
{[}, \\
-
\end{array}
\right|
\left.
\begin{array}{l}
{]}, \\
{]}
\end{array}
\right|
\left.
\begin{array}{l}
|, \\
{]}
\end{array}
\right) \\
\s{B}_n & \to & \left(
\begin{array}{l}
-, \\
-
\end{array}
\right|
\left.
\begin{array}{l}
+, \\
+
\end{array}
\right|
\left.
\begin{array}{l}
{[}, \\
{[}
\end{array}
\right|
\left.
\begin{array}{l}
{]}, \\
{]}
\end{array}
\right|
\left.
\begin{array}{l}
|, \\
|
\end{array}
\right) \\
\end{array}
\]
continuing with
\[
\begin{array}{rcccccc}
\s{NBR} & \to & A & A & - & - & - \\
\s{RLE} & \to & A & A & - & {[} & - \\
\s{RRE} & \to & A & A & - & {]} & - \\
\s{BLE} & \to & A & A & A & - & {[} \\
\s{BRE} & \to & A & A & A & - & {]} \\
\s{RB} & \to & A & A & A & {]} & {[} \\
\s{BR} & \to & A & A & A & {[} & {]} \\
\s{RED} & \to & A & A & - & + & - \\
\s{BAS} & \to & A & A & A & - & + \\
\end{array}
\]
In cases where the reduplicant precedes the base, the reduplication rules will appear as follows:
\[
\begin{array}{rcl}
\s{Rd1} & \to & \left(
\begin{array}{lllll} 
\s{BDR}_0& \s{UR}& \s{MRD}& \s{RLE}& \s{Rd2}_0, \\
\s{BDL}_0& \s{UR}& \s{BDR}_1& \s{RB}& \s{Rd2}_1 \\
\s{SSR}& \s{UR}& \s{BDL}_1& \s{BRE} \\
\end{array}
\right) \\
\s{Rd2} & \to & \left(
\begin{array}{llll}
B_0& \s{UR}& \s{MRD}& \s{RED}, \\
\s{SSR}& \s{UR}& B_1& \s{BAS}
\end{array}
\right) \\
        & |  & \left(
\begin{array}{lllll}
\s{Rd2}_0& B_0& \s{UR}& \s{MRD}& \s{RED}, \\
\s{Rd2}_1& \s{SSR}& \s{UR}& B_1& \s{BAS}
\end{array}
\right) \\
\end{array}
\]

Otherwise, where the base precedes the reduplicant, the rules will appear as
follows:
\[
\begin{array}{rcl}
\s{Rd1} & \to & \left(
\begin{array}{lllll} 
\s{SSR}& \s{UR}& \s{BDR}_1& \s{BLE}& \s{Rd2}_0, \\
\s{BDR}_0& \s{UR}& \s{BDL}_1& \s{BR}& \s{Rd2}_1 \\
\s{BDL}_0& \s{UR}& \s{MRD}& \s{RRE} \\
\end{array}
\right) \\
\s{Rd2} & \to & \left(
\begin{array}{llll}
\s{SSR}& \s{UR}& B_1& \s{BAS}, \\
B_0& \s{UR}& \s{MRD}& \s{RED}
\end{array}
\right) \\
        & |  & \left(
\begin{array}{lllll}
\s{Rd2}_0 & \s{SSR}& \s{UR}& B_1& \s{BAS}, \\
\s{Rd2}_1 & B_0& \s{UR}& \s{MRD}& \s{RED}
\end{array}
\right) \\
\end{array}
\]
\end{footnotesize}

\caption{Reduplication Grammar\label{fig:grammar}}

\end{figure}

\section{The Earley Algorithm}

\begin{figure}[htb]
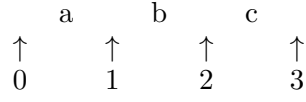

\begin{tabular}{cccccccc}
    & a &      & b &     & c & \\
$\uparrow$ &   & $\uparrow$ &   & $\uparrow$ &   & $\uparrow$ \\
0   &   & 1    &   & 2   &   & 3 \\
\end{tabular}
\caption{\label{fig:numbering}Numbering of string positions in the
  string ``abc''}
\end{figure}

The Earley algorithm is an efficient chart parsing method.  Chart
parsing can be seen as a method for taking the intersection of a
string or FSM with a CFG (later, an MCFG).  Here we take a CFG as a
4-tuple $\langle V,N,P,S\rangle$ where $V$ represents the set of terminals in the
grammar, $N$ represents the set of non-terminals, $P$ represents the
set of productions, and $S\in N$ is the start symbol.  In the
definitions to follow, $\alpha$, $\beta$, and $\gamma$ represent arbitrary members
of $(V\cup N)^{*}$, $A$ and $C$ represent arbitrary members of $N$, $a$
and $b$ represent arbitrary members of $V$, $p$ represents an
arbitrary member of $P$, and the indices $i$, $j$, and $k$ represent
positions within the input string to be parsed, numbered as in
Figure~\ref{fig:numbering}.

In the standard definition, a member of the chart is a 3-tuple
$(i,C\to\alpha\bullet\beta,j)$, where $i$ represents the position at the beginning of
the input string covered by $\alpha$ and $j$ represents the position at the
end of the covered portion of the string.  The parsing operation in
the standard definition, which parses a single input string, is
defined as a closure via the following three inference rules of a
chart initially consisting of $(0,S\to\bullet\alpha,0)$:
\begin{description}
\item [predict:]\( \frac{(i,C\to \alpha \bullet A\beta ,j)}{(j,A\to \bullet \gamma ,j)} \)
if \( A\to \gamma \in P \) (if \( \gamma  \) begins with a terminal,
that terminal must be the symbol at position $j$ in the input string)
\item [scan:]\( \frac{(i,C\to \alpha \bullet a\beta ,j)}{(i,C\to \alpha a\bullet \beta ,j+1)} \)
if \( a\) is the symbol after $j$
\item [complete:]\( \frac{(i,C\to \alpha \bullet A\beta ,j)\; (j,A\to \gamma \bullet ,k)}{(i,C\to \alpha A\bullet \beta ,k)} \)
\end{description}
The input string is recognized if the chart contains an element
$(0,S\to\alpha\bullet,n)$, where $n$ is the final position of the input string.

\section{Extending Earley}

The algorithm presented so far just checks to see whether a particular
string exists in a grammar. In order for it to be useful for our
purposes, the following extensions must be made:
\begin{enumerate}
\item Intersection with an FSM, not just a string
\item Recovery of intersection grammar
\item Weights (intersection should allow lowest-weight derivations only)
\item MCFGs
\end{enumerate}

\subsection{Intersection with an FSM}

To modify the algorithm to intersect a grammar with an FSM, we replace
the input string with an FSM, and change our definition of a chart
entry.  Now, a chart entry is a 3-tuple $(i,C \to \alpha\bullet\beta,j)$, where $i$
represents the first FSM state covered by $\alpha$ and $j$ represents the
last FSM state covered.  We define an FSM here as a 5-tuple
$\langle Q,\Sigma,s,F,M\rangle$, where $Q$ is the set of states in the FSM, $\Sigma$ is the
label alphabet for the FSM (for our purposes $\Sigma$ is always the same as
$V$ for all grammars in use), $s\in Q$ is the start state, $F\subseteq Q$ is the
set of final states of the FSM, and $M$ is a set of 3-tuples
$(i,a,j)$, which represent transitions from state $i$ to state $j$
with label $a$.  Given these redefinitions we can then just modify the
scan rule:

\begin{description}
\item [scan:]\( \frac{(i,C\to \alpha \bullet a\beta ,j)}{(i,C\to \alpha a\bullet \beta ,j+1)} \)
if \( (j,a,k)\in M \), where \( M \) is the input FSM.
\end{description}
and the predict rule in the obvious way:

\begin{description}
\item [predict:]\( \frac{(i,C\to \alpha \bullet A\beta ,j)}{(j,A\to \bullet \gamma ,j)} \)
if \( A\to \gamma \in P \) (if \( \gamma  \) is of the form \( a\; \gamma ^{\prime } \),
\( (j,a,k)\in M \) must hold as well)
\end{description}

Note that the initial entry in the chart is now $(s,S\to\bullet\alpha,s)$.

\subsection{Grammar  Recovery}

It is possible to recover the output of intersection by increasing
slightly what is in the chart.  In particular, for every item on the
chart, we note how it got there (just the last step).  Each item on
the chart may be referred to by its column number \( C \) and its
position \( N \) within that column.  We annotate only items produced
by scan and complete steps, as follows:
\begin{itemize}
\item \( sC/N \)
\item \( cC_{1}/N_{1};C_{2}/N_{2} \)
\end{itemize}
where $C_1/N_1$ refers to the $(j,A\to \gamma \bullet ,k)$ item from the complete
step, and $C_2/N_2$ refers to the $(i,C\to \alpha \bullet A\beta ,j)$ item. A chart
item is thus now a 4-tuple $(i,C\to\alpha\bullet\beta,j,H)$, where $H$ is a set of
history items of the type described here, one for each scan or
complete step that put the item there.

Recovery of a grammar then starts from the ``success items,'' that is
items in the chart that begin in state 1 and end with a final state
and represent a production from the start symbol of the grammar, with
the Earley position dot at the end of the production.  We then move
from right to left within those productions, filling in the state pairs for
each constituent we pass, and tracing through their productions as
well.  Whenever we get to the left side of a production, we output
it.  The exact algorithm is as follows:
\begin{description}
\item[GrammarRecovery(\emph{chart})] \hfill
\begin{algorithmic}
\STATE $\s{queue} \gets []$
\FORALL{$\s{success items}\: (s, S\to\gamma \bullet, f\in F,H_0)$ at $(C, N)$}
  \STATE queue up $(C, N)$ onto \emph{queue}
  \WHILE{\emph{queue} not empty}
    \STATE $(C, N) \gets $ dequeue from \emph{queue}
    \STATE \emph{item} $\gets$ item at $(C, N)$: $(i, A\to\alpha \bullet, j,H_1)$
    \STATE \emph{pos} $\gets$ pos. of $\bullet$ in \emph{item}
    \STATE \emph{RHSs} $\gets$ \textbf{GetRHSs(}[[]], \emph{item, pos, queue}\textbf{)}
    \FORALL{\emph{RHS} $\in$ \emph{RHSs}}
      \STATE output ``$A(i,j)\to\s{RHS}$''
    \ENDFOR
  \ENDWHILE
\ENDFOR
\end{algorithmic}
\item[GetRHSs(\emph{rhss, item, pos, queue})] \hfill
\begin{flushleft}
\begin{algorithmic}
\IF{$\s{pos} = 0$}
\STATE return \emph{rhss}
\ENDIF
\STATE \emph{new\_rhss} $\gets []$
\FORALL{history path components \emph{hitem} of \emph{item}}
  \STATE $\s{rhss}^\prime \gets$ copy \emph{rhss}
  \STATE \textbf{extend(}$\s{rhss}^\prime$, \emph{hitem, pos, queue}\textbf{)}
  \STATE add $\s{rhss}^\prime$ to \emph{new\_rhss}
\ENDFOR
\STATE return \emph{new\_rhss}
\end{algorithmic}
\end{flushleft}
\item[extend(\emph{rhss, hitem, pos, queue})] \hfill
\begin{flushleft}
\begin{algorithmic}
\IF{\emph{hitem} = s$(C,N)$}
  \STATE prepend scanned symbol to each \emph{rhs} $\in$ \emph{rhss}
  \STATE \emph{prev} $\gets$ item at $(C,N)$
\ELSIF{\emph{hitem} = c$(C_1/N_1;C_2/N_2)$}
  \STATE $(i,A\to\gamma\bullet,j,H) \gets$ item at $(C_1,N_1)$
  \STATE prepend $A(i,j)$ to each \emph{rhs} $\in$ \emph{rhss}
  \STATE enter $(C_1,N_1)$ into \emph{queue}
  \STATE \emph{prev} $\gets$ item at $(C_2,N_2)$
\ENDIF
\STATE return \textbf{GetRHSs(}\emph{rhss, item, pos$-1$, queue}\textbf{)} 
\end{algorithmic}
\end{flushleft}
\end{description}

\subsection{Weights}

The basic idea for handling weights is an adaption from the Viterbi algorithm,
as used for chart parsing of probabilistic grammars. Basically, we reduce the
grammar to allow only the lowest-weight derivations from each new category.

\paragraph{Implementation:}

Each chart item has an associated weight, computed as follows:
\begin{description}
\item [predict:]weight of the predicted rule \( A\to \gamma  \)
\item [scan:]sum of the weight of the item scanned from and the weight of the FSM
edge scanned across.
\item [complete:]sum of the weights of the two items involved
\end{description}
We build new chart items whenever permitted by the rules given in
previous sections, assigning weights to them by the above
considerations. If no equivalent item (equivalence ignores weight and
path to the item) is in the chart, we add the item.  If an equivalent
item is in the chart, there are three possible actions, according to
the weight of the new item:
\begin{enumerate}
\item Higher than the old item: do nothing (don't add the new path).
\item Lower than the old item: remove all other paths to the item, add
  this path to the item. Adjust weights of all items built from this
  one downward.
\item Same as the old item: add the new path to the item.
\end{enumerate}
A chart item is thus now a 5-tuple $(w,i,C\to\alpha\bullet\beta,j,H)$, where $w$
represents a weight, and all the other items are as before.

\subsection{MCFGs}

To extend the Earley algorithm to MCFGs, we first reduce the
chart-building part of the Earley algorithm for MCFGs to the
already-worked out algorithm for CFGs by converting the MCFG into a
(not-equivalent) CFG.  We then modify the grammar-recovery step to
convert the CFG produced into an MCFG, verifying that the MCFG
produced is a proper one.

\subsubsection{Adjustments to the Chart-Building Algorithm:}

First, we treat each part of the rule as a separate rule, and use the regular algorithm.
Thus, B\( \to  \)\( \left( \begin{array}{c}
0\\
1
\end{array}\right)  \) becomes B\( _{0} \)\( \to 0 \) and B\( _{1} \)\(
\to 1 \).
Having separated a single rule such as $C\to(\alpha,\beta)$ into two parts
$C_0\to\alpha$ and $C_1\to\beta$, we need to keep track, when building the chart
and after, of which rule in the associated MCFG each chart item refers
to.  These annotations will be useful in Grammar Recovery (something
like $C_0\to\alpha$ can only be combined with $C_1\to\beta$ if they both come from
the same MCFG rule).  Thus, a chart item is a 6-tuple
$(r,w,i,C\to\alpha\bullet\beta,j,H)$, where $r$ is the rule number from the original
MCFG to which the production $C\to\alpha\bullet\beta$ corresponds, and all the others
are as before.

\subsubsection{Adjustments to Grammar Recovery}

As before, followed by a final combinatory and checking step:
\begin{algorithmic}
\FORALL{non-terminals $A$ with arity $n$}
  \FORALL{possible combinations
    $A_0(i,j)\to\gamma_0, A_1(k,l)\to\gamma_1,\ldots, A_n(m,n)\to\gamma_n$}
    \IF{the MCFG condition applies to the combination}
      \STATE output $A(i,j)(k,l)\ldots(m,n)\to(\gamma_0,\gamma_1,\ldots,\gamma_n)$
    \ENDIF
  \ENDFOR
\ENDFOR
\end{algorithmic}
where the MCFG condition is as follows:
\begin{quote}
  All $\gamma_i$ on the right hand side of the combination must be derived
  from the same rule in the original set of rules and their yields
  must not overlap each other in the FSM.
\end{quote}
Given the way the chart-parsing and recovery algorithms work, the MCFG
condition will be satisfied if we simply check that all the elements
of the combination come from the same rule in the original MCFG.  This
will result in some invalid rules in the output grammar, but this
simple check guarantees that these rules will be such that they will
be unable to participate in derivations, since their right-hand sides
will refer to categories that do not head any productions.

\subsection{Example}

As an example, let's take the simple reduplication grammar from
before:
\[
\begin{array}{crcl}
(1) & S & \to & A_0 \: A_1 \\
(2) & A & \to & (1, 1) \\
(3) &   & | & (0, 0) \\
(4) &   & | & (0 \: A_0, 0 \: A_1) \\
(5) &   & | & (1 \: A_0, 1 \: A_1)
\end{array}
\]
and intersect it with the machine
\begin{quote}
{\par\centering \includegraphics{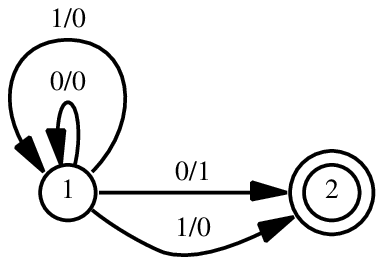} \par}
\end{quote}
This machine generates the set of strings $\{0|1\}^{+}$, but weights all
strings ending with $0$.  

The corresponding CFG-grammar used for the chart-building step is as
follows:
\[
\begin{array}{crcl}
(1) & S & \to & A_0 \: A_1 \\
(2) & A_0 & \to & 1 \\
    & A_1 & \to & 1 \\
(3) & A_0 & \to & 0 \\
    & A_1 & \to & 0 \\
(4) & A_0 & \to & 0 \: A_0 \\
    & A_1 & \to & 0 \: A_1 \\
(5) & A_0 & \to & 1 \: A_0 \\
    & A_1 & \to & 1 \: A_1 \\
\end{array}
\]

The chart produced by the chart-building part of the algorithm is as
follows:
\begin{center}
Column 1 ($j=1$, $i=1$)
\end{center}
\begin{footnotesize}
\[
\begin{array}{r|cccc}
 \# & r & w & \in P         & H \\ \hline
 0 & 1 & 0 & S\to\bullet A_0\:A_1 & \emptyset \\
 1 & 2 & 0 & A_0\to\bullet1     & \emptyset \\
 2 & 3 & 0 & A_0\to\bullet0     & \emptyset \\
 3 & 4 & 0 & A_0\to\bullet0\:A_0 & \emptyset \\
 4 & 5 & 0 & A_0\to\bullet1\:A_0 & \emptyset \\
 5 & 5 & 0 & A_0\to1\bullet A_0   & \{s1/4\} \\
 6 & 4 & 0 & A_0\to0\bullet A_0   & \{s1/3\} \\
 7 & 3 & 0 & A_0\to0\bullet     & \{s1/2\} \\
 8 & 1 & 0 & S\to A_0\bullet A_1   & \\ 
   & \multicolumn{4}{l}{\{c1/7;1/0, c1/10;1/0, c1/9;1/0,
c1/22;1/0\}} \\
 9 & 5 & 0 & A_0\to1\:A_0\bullet & \\
   & \multicolumn{4}{l}{\{c1/7;1/5, c1/10;1/5, c1/9;1/5, c1/22;1/5\}} \\ 
10 & 4 & 0 & A_0\to0\:A_0\bullet & \\
   & \multicolumn{4}{l}{\{c1/7;1/6, c1/10;1/6, c1/9;1/6, c1/22;1/6\}}
   \\
11 & 2 & 0 & A_1\to\bullet1     & \emptyset \\
12 & 3 & 0 & A_1\to\bullet0     & \emptyset \\
13 & 4 & 0 & A_1\to\bullet0\:A_1 & \emptyset \\
14 & 5 & 0 & A_1\to\bullet1\:A_1 & \emptyset \\
15 & 5 & 0 & A_1\to1\bullet A_1   & \{s1/14\} \\
16 & 4 & 0 & A_1\to0\bullet A_1   & \{s1/13\} \\
17 & 3 & 0 & A_1\to0\bullet     & \{s1/12\} \\
18 & 1 & 0 & S\to A_0\:A_1\bullet & \\
   & \multicolumn{4}{l}{\{c1/17;1/8, c1/20;1/8, c1/19;1/8, c1/21;1/8\}}
   \\
19 & 5 & 0 & A_1\to1\:A_1\bullet & \\
   & \multicolumn{4}{l}{\{c1/17;1/14, c1/20;1/14, c1/19;1/14, c1/21;1/14\}}
   \\
20 & 4 & 0 & A_1\to0\:A_1\bullet & \\
   & \multicolumn{4}{l}{\{c1/17;1/13, c1/20;1/13, c1/19;1/13, c1/21;1/13\}}
   \\
21 & 2 & 0 & A_1\to1\bullet     & \{s1/11\} \\
22 & 2 & 0 & A_0\to1\bullet     & \{s1/1\} \\
\end{array}
\]
\end{footnotesize}
\begin{center}
Column 2 ($j=2$, $i=1$)
\end{center}
\begin{footnotesize}
\[
\begin{array}{r|cccc}
 \# & r & w & \in P         & H \\ \hline
 0 & 5 & 0 & A_0\to1\bullet A_0   & \{s1/4\} \\
 1 & 4 & 1 & A_0\to0\bullet A_0   & \{s1/3\} \\
 2 & 3 & 1 & A_0\to0\bullet     & \{s1/2\} \\
 3 & 1 & 0 & S\to A_0\bullet A_1   & \\
   & \multicolumn{4}{l}{\{c2/13;1/0, c2/5;1/0, c2/4;1/0\}} \\
 4 & 5 & 0 & A_0\to1\:A_0\bullet & \\
   & \multicolumn{4}{l}{\{c2/13;1/5, c2/5;1/5, c2/4;1/5\}} \\
 5 & 4 & 0 & A_0\to0\:A_0\bullet & \\
   & \multicolumn{4}{l}{\{c2/13;1/6, c2/5;1/6, c2/4;1/6\}} \\
 6 & 5 & 0 & A_1\to1\bullet A_1   & \{s1/14\} \\
 7 & 4 & 1 & A_1\to0\bullet A_1   & \{s1/13\} \\
 8 & 3 & 1 & A_1\to0\bullet     & \{s1/12\} \\
 9 & 1 & 0 & S\to A_0\:A_1\bullet & \\
   & \multicolumn{4}{l}{\{c2/12;1/8, c2/11;1/8, c2/10;1/8\}} \\
10 & 5 & 0 & A_1\to1\:A_1\bullet & \\
   & \multicolumn{4}{l}{\{c2/12;1/15, c2/11;1/15, c2/10;1/15\}} \\
11 & 4 & 0 & A_1\to0\:A_1\bullet & \\
   & \multicolumn{4}{l}{\{c2/12;1/16, c2/11;1/16, c2/10;1/16\}} \\
12 & 2 & 0 & A_1\to1\bullet     & \{s1/11\} \\
13 & 2 & 0 & A_0\to1\bullet     & \{s1/1\} \\
\end{array}
\]
\end{footnotesize}
In this chart the items with an empty history list were entered by
prediction steps.  The ``success item'' for this grammar is then item
(2,9): $(r=1,w=0,i=1,p=S\to A_0\:A_1\bullet,j=2,H=\{c2/12;1/8, c2/11;1/8,
  c2/10;1/8\})$, so begin there:
\[
  S(1,2) \to A_0 \: A_1
\]
We then queue up (2,12), (2,11), and (2,10), noting that for all of
these the states for $A_1$ are (1,2), and we move to item (1,8):
$(r=1,w=0,i=1,p=S\to A_0\bullet A_1,j=1,H=\{c1/7;1/0, c1/10;1/0, c1/9;1/0,
c1/22;1/0\})$.  Here we queue up (1,7), (1,10), (1,9), and (1,22), noting that
for all of these the states for $A_0$ are (1,1).  Moving to (1,0), we
note that we are done, and we thus output a complete rule:
\[
(r1) \: S(1,2) \to A_0(1,1) \: A_1(1,2).
\]
We then encounter (2,12) on the queue:
$(r=2,w=0,i=1,p=A_1\to1\bullet,j=2,H=\{s1/11\})$, which can be output with no
further ado:
\[
(r2) \: A_1(1,2) \to 1
\]
Moving to item (2,11) $(r=4,w=0,i=1,p=A_1\to0\:A_1\bullet,j=2,H=\{c2/12;1/16,
c2/11;1/16, c2/10;1/16\})$ we don't need to queue anything, and we can
see that the output will be:
\[
(r4) \: A_1(1,2) \to 0 \: A_1(1,2)
\]
Item (2,10) is $(r=5, w=0, i=1,p=A_1\to1\:A_1\bullet, j=2, H=\{c2/12;1/15,
  c2/11;1/15, c2/10;1/15\})$, so we output
\[
(r5) \: A_1(1,2) \to 1 \: A_1(1,2)
\]
We now move on to item (1,7): $(r=3,w=0,i=1,p=A_0\to0\bullet,j=1,H=\{s1/2\})$,
which we output as
\[
(r3) \: A_0(1,1) \to 0.
\]
Item (1,10) is $(r=4,w=0,i=1,p=A_0\to0\:A_0\bullet,j=1,H=\{c1/7;1/6, c1/10;1/6,
c1/9;1/6, c1/22;1/6\})$.  In dealing with this we need to queue
nothing, and we output:
\[
(r4) \: A_0(1,1) \to 0 \: A_0(1,1)
\]
Moving to (1,9), which is $(r=5,w=0,i=1,p=A_0\to1\:A_0\bullet,j=1,H=\{c1/7;1/5,
c1/10;1/5, c1/9;1/5, c1/22;1/5\})$, we queue nothing and output
\[
(r5) \: A_0(1,1) \to 1 \: A_0(1,1)
\]
Finally we get to (1,22): $(r=2,w=0,i=1,p=A_0\to1\bullet,j=1,H=\{s1/1\})$, which
gets output as
\[
(r2) \: A_0(1,1) \to 1
\]
Collecting these together (for category $A$), we get the following
pairings:
\begin{scriptsize}
\[
\left(
\begin{array}{c|rcl|rcl}
(r2) & A_0(1,1) & \to & 1 & A_1(1,2) & \to & 1 \\
(r3) & A_0(1,1) & \to & 0 & & & \\
(r4) & A_0(1,1) & \to & 0 \: A_0(1,1) & A_1(1,2) & \to & 0 \: A_1(1,2) \\
(r5) & A_0(1,1) & \to & 1 \: A_0(1,1) & A_1(1,2) & \to & 1 \: A_1(1,2) \\
\end{array}
\right)
\]
\end{scriptsize}
Note that the ``pair'' for (r3) has no second member, so nothing will
be output for it.  Combining the
compatible rules, we get the following grammar:
\[
\begin{array}{rcl}
S(1,2) & \to & A(1,1)(1,2)_0 \: A(1,1)(1,2)_1 \\
A(1,1)(1,2) & \to & (1,  1) \\
            & |  & (0 \: A_0(1,1)(1,2), 0 \: A_1(1,1)(1,2)) \\
            & |  & (1 \: A_0(1,1)(1,2), 1 \: A_1(1,1)(1,2)) \\
\end{array}
\]
which is equivalent to the grammar:
\[
\begin{array}{rcl}
S & \to & A_0 \: A_1 \\
A & \to & (1, 1) \\
  & | & (0 \: A_0, 0 \: A_1) \\
  & | & (1 \: A_0, 1 \: A_1) \\
\end{array}
\]
This grammar indeed represents the best outputs from the intersection---all
reduplicating forms which end in a 1.

\bibliography{abbrevs,full}
\end{document}